\documentclass[10pt,twocolumn,letterpaper]{article}

\usepackage[pagenumbers]{cvpr} 

\usepackage{graphicx}
\usepackage{amsmath}
\usepackage{amssymb}
\usepackage{booktabs}

\usepackage{listings}

\usepackage[ruled,vlined]{algorithm2e}
\usepackage{amsthm}

\usepackage[breaklinks=true,colorlinks,bookmarks=false]{hyperref}

\usepackage[symbol]{footmisc}
\usepackage{diagbox}
\usepackage{multirow}

\newtheorem{proposition}{Proposition}
\newtheorem{lemma}{Lemma}

\newtheorem{assumption}{Assumption}

\newtheorem*{innerthm}{\innerthmname}
\newcommand{\innerthmname}{}
\newenvironment{statement}[1]
 {\renewcommand{\innerthmname}{#1}\innerthm}
 {\endinnerthm}
\theoremstyle{definition}

\newcommand*{\affaddr}[1]{#1} 

\usepackage[capitalize]{cleveref}
\crefname{section}{Sec.}{Secs.}
\Crefname{section}{Section}{Sections}
\Crefname{table}{Table}{Tables}
\crefname{table}{Tab.}{Tabs.}

\def\cvprPaperID{*****} 
\def\confName{CVPR}
\def\confYear{2022}

\newcommand{\LL}{\mathcal{L}}
\newcommand{\kld}{D_{KL}}
\newcommand{\red}[1]{#1}

\begin{document}

\title{Self-Knowledge Distillation via Dropout}



\author{%
Hyoje Lee, Yeachan Park, Hyun Seo, Myungjoo Kang\\
\affaddr{Seoul National University}\\
{\tt\small \{hyoje42, ychpark, hseo0618, mkang\}@snu.ac.kr}
}

\maketitle

\begin{abstract}
To boost the performance, deep neural networks require deeper or wider network structures that involve massive computational and memory costs. To alleviate this issue, the \textit{self-knowledge distillation} method regularizes the model by distilling the internal knowledge of the model itself. Conventional self-knowledge distillation methods require additional trainable parameters or are dependent on the data. In this paper, we propose a simple and effective self-knowledge distillation method using a dropout (\textit{SD-Dropout}). \textit{SD-Dropout} distills the posterior distributions of multiple models through a dropout sampling. Our method does not require any additional trainable modules, does not rely on data, and requires only simple operations. Furthermore, this simple method can be easily combined with various self-knowledge distillation approaches. 
We provide a theoretical and experimental analysis of the effect of forward and reverse KL-divergences in our work.
Extensive experiments on various vision tasks, i.e., image classification, object detection, and distribution shift, demonstrate that the proposed method can effectively improve the generalization of a single network.
Further experiments show that the proposed method also improves  calibration performance, adversarial robustness, and out-of-distribution detection ability.
\end{abstract}


\section{Introduction}

Deep neural networks (DNN) have achieved a state-of-the-art performance in many domains, including image classification, object detection, and segmentation \cite{vggpaper, resnet, maskrcnn}.
In designing models that are deeper and more complex for a higher performance, model compression is essential in delivering a deep learning model for practical application.
To develope lightweight models, many previous attempts have been made, including an efficient architecture \cite{howard2017mobilenets, tan2019efficientnet}, model quantization \cite{quantization}, pruning \cite{pruning}, and knowledge distillation \cite{hinton2015distilling}.

Although knowledge distillation is a popular DNN compression method, 
conventional offline knowledge distillation methods have several limitations.
To distill the knowledge from a teacher network to a student network, two main steps are required.
First, we train a large teacher network, followed by a student network using distillation.
Fully training the teacher model with massive datasets requires considerable effort.
Second, it is difficult to search for an appropriate teacher model that corresponds to the target student model.
In addition, The common belief in traditional knowledge distillation is to expect a larger or more accurate teacher network to be a good teacher. 
However, a teacher network with a deeper structure and higher accuracy does not guarantee the improved performance of the student network \cite{cho2019efficacy, Yuan_Revisiting_2020_CVPR}.

Self-knowledge distillation is a solution to these limitations. In a self-knowledge distillation, a teacher network becomes a student network itself. Knowledge is efficiently distilled in a single training process in a single model without the guidance of other external models. Several self-distillation methods have been proposed \cite{dml, xu2019data-distortion, byot, cskd}.  However, these methods also have the following drawback: 1) Some methods require subnetworks with additional parameters. 2) Some methods request additional ground-truth label information, which means that the model depends on the class distribution of the training datasets.


Inspired by these observations, we propose a simple self-knowledge distillation using a dropout (SD-Dropout).
Our method generates ensemble models with identical architecture, but different weights through a dropout sampling.
After all feature extraction layers, we sample the global feature vector to obtain two different features with different perspectives. These two feature vectors are then passed to the last fully connected layer, and two different posterior distributions are generated.
We then match these posterior distributions using Kullback-Leibler divergence (KL-divergence).
Dark knowledge between these two internal models can improve their performance through knowledge distillation \cite{hinton2015distilling}.

The proposed method does not require any additional parameters nor does it require additional label information. For these reasons, SD-Dropout is computationally more efficient than other self-distillation methods.
Furthermore, the SD-Dropout method is model-agnostic and method-agnostic, meaning it can be easily implemented with various backbone models and other self-distillation methods.

For more effective distillation, we consider the way to use KL-divergence. In most of the other methods, the gradient of the reference distribution in the KL-divergence is not propagated through model parameters (forward direction of KL divergence). 
We propose a new approach of utilizing the reverse direction of KL divergence by propagating the gradient flow of the reference distribution.  We theoretically demonstrate that the gradient of the reference distribution is greater than that of the other distribution, and we empirically verify the effectiveness of using the gradient of both distributions in the KL-divergence.

We conduct extensive experiments to verify the effectiveness and generalization of our method on various image classification tasks, \red{CIFAR-100 \cite{cifar}, CUB-200-2011 \cite{cub}, and Stanford Dog \cite{stanforddog}. Also, we verify that our method works well on the large-scale dataset, ImageNet \cite{ImageNet}, and the distribution shift dataset CIFAR-C dataset \cite{hendrycks2019benchmarking}. We also examine that the proposed method improves the performance in the object detection on the MS COCO dataset \cite{lin2014microsoft}.} In addition, when acquiring two different sampled feature vectors, only a dropout layer is used after obtaining the feature vectors; thus, the proposed method can be easily applied to any network architecture structure and any knowledge distillation methods. 

Experimental results demonstrate that our simple and effective regularization method improves the performance of various model architectures, \red{ResNet \cite{resnet} and DenseNet \cite{huang2017densely},} and is in good agreement with other knowledge distillation methods \cite{dml, xu2019data-distortion, byot, cskd}. Furthermore, our experiments show that our method improves the calibration performance, \red{adversarial robustness, and out-of-distribution detection ability}.

Our contributions are summarized as follows:

\begin{itemize}
\item We present a simple self-knowledge distillation methodology using dropout techniques.
\item Our self-knowledge distillation method can collaborate easily with other knowledge distillation methods.
\item We describe experimental observations regarding the forward and \red{reverse} KL-divergence commonly used in knowledge distillation.
\item Extensive experiments demonstrate the effectiveness of our methodology.
\end{itemize}

\section{Related Work}

\red{Knowledge distillation \cite{hinton2015distilling} is a learning method for transferring knowledge from a large and complex network (known as a teacher network) into a small and simple network (known as a student network). 
A number of variants have been proposed, inspired by the original method of knowledge distillation, such as the previously mentioned teacher-student framework \cite{passalis2018elearning_prob_KD, heo2019knowledge_adv_samples, mirzadeh2020improved_TA_KD}.}

\begin{figure*}[t]
\begin{center}
   \includegraphics[width=0.9\textwidth]{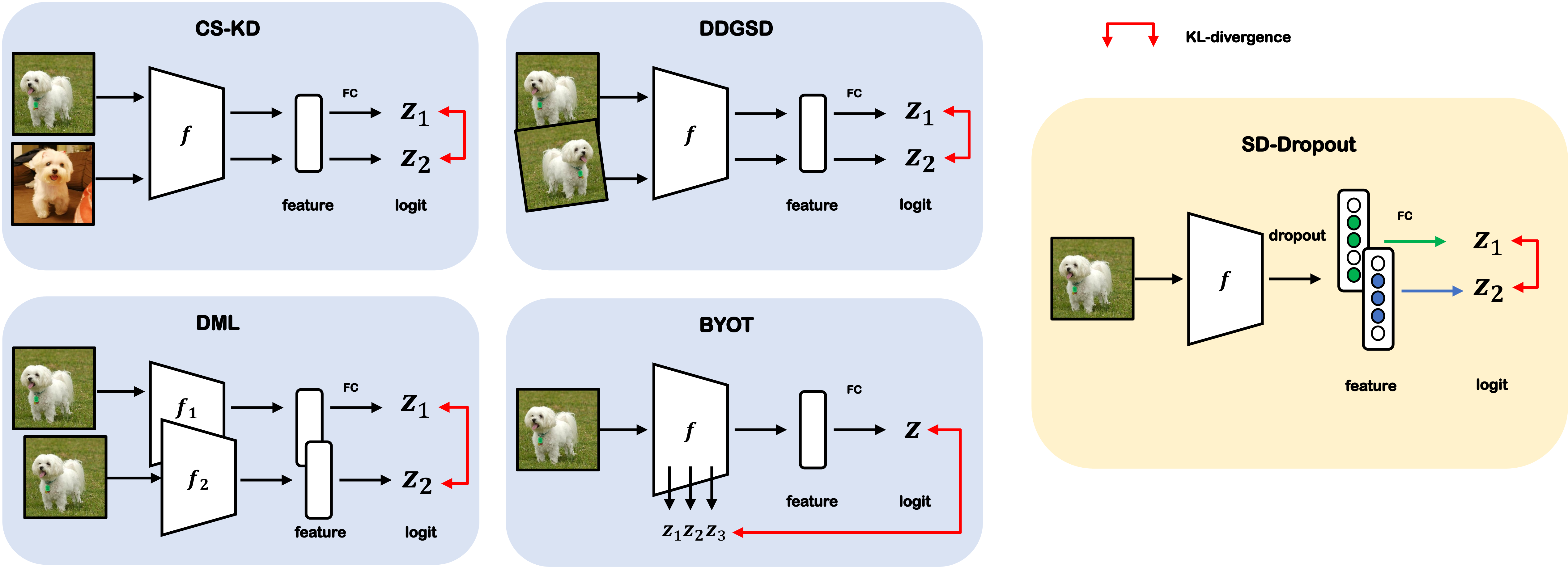}
\end{center}
   \caption{\red{Design of the Self-Knowledge Distillation Methods. \textit{(Left Blue Background)} The designs of existing self-knowledge distillation method. \textit{(Right Yellow Background)} Our method, SD-Dropout, samples the global feature vector to obtain different features using dropout. These two different features induce different two posterior distributions. The knowledge between the two distributions is distilled to each other. Red arrows indicate distillation using KL-divergence.} }
\label{fig:self KD methods}
\end{figure*}

\red{On the other hand, research that breaks away from the aforementioned teacher-student framework has been proposed. Deep Mutual Learning (DML) \cite{dml} is a distillation method in which a teacher network and a student network distill the knowledge from each other. Data-Distortion Guided Self-Distillation (DDGSD) \cite{xu2019data-distortion} distills knowledge between different distorted data. Be Your Own Teacher (BYOT) \cite{byot} distills knowledge between its deeper and shallow layers. Class-wise self-knowledge distillation (CS-KD) \cite{cskd} matches the posterior distributions of a model between intra-class instances. We visualize these prior methods in diagram forms in Figure \ref{fig:self KD methods}.}

\red{These methods have attempted to distill the knowledge from a network within itself. Note that prior methods require subnetworks with additional parameters or request additional label information. Also, some research requires an additional procedure to distort the input data. 
We remark that these methods are computationally
expensive by training additional networks from scratch or traversing two instances from the beginning to the end of a network,
and these points are different from ours.}

\red{Similar to our work, several semi-supervised and self-supervised learning methods have been investigated. Several studies attempt to solve the semi-supervised task using past selves as teacher networks \cite{laine2016temporal, tarvainen2017mean}. Our method and the self-supervised literature  \cite{he2020momentum,chen2020simple,grill2020bootstrap, chen2021simple_siamese, gidaris2021obow, zbontar2021barlow} have a similar idea of comparing two representations, however, ours does not require augmentations of the input data and additional modules with parameters. The idea of distillation from the ensemble of model for uncertainty estimation is also investigated \cite{li2018reducing,englesson2019efficient,malinin2019ensemble,malinin2019reverse_kl}. }

\red{We remark that there are several works on knowledge distillation related to dropout. Specifically, there are several works that distill the output from Monte Carlo dropout techniques \cite{bulo2016drop_distill, gurau2018drop_dist_for_efficiently}. These methods investigate distilling the knowledge obtained from averaging Monte Carlo samples into the student model to distill uncertainty. However, these approaches attempt to train the model by mimicking  the ensembled prediction obtained by multiple Monte Carlo sampling, we focus on distilling knowledge between internal features obtained from dropout sampling within the model.}

\section{Self Distillation via Dropout}

Throughout this study, we focus on supervised classification tasks. We denote $\mathbf{x} \in \mathcal{X}$ as the input data and $y \in \mathcal{Y}=\{1, 2, ..., N\}$ as its ground-truth label class. Let $f(\mathbf{x})$ be a global feature vector of the input data $\mathbf{x}$, and let $h(\cdot)$ be the last fully connected layer in a network. Now, we define $\mathbf{z}=\mathcal{M}_{\theta}(\mathbf{x})=h(f(\mathbf{x}))$ as the logit of the output layer, where $\mathcal{M}_{\theta}$ is the neural network parametrized by $\theta$. In classification tasks, neural networks typically use a \textit{softmax} classifier to produce class posterior probability. Thus, we can consider that the posterior probability of class $i$ is as follows:
\begin{equation}
     p(y=i | \mathbf{x} ; \theta, T) = \frac{exp(z^{i}/T)}{\sum_{j}^{N}{exp(z^{j}/T)}},
\end{equation}

where $z^{i}$ as the logit of class $i$ and $T>0$ is the temperature, which is usually set to 1. In knowledge distillation, the temperature $T$ is set to greater than 1.

\subsection{Method Formulation}

In this section, we introduce a new self-knowledge distillation method called SD-Dropout. We use the dropout layer after all feature extraction layers. We define
\begin{equation}
    \mathcal{M}^{\mathbf{u}}_\theta(\mathbf{x}) = h(\mathbf{u} \odot f(\mathbf{x}))  \qquad \textrm{where $u^{j}$ $\sim$ Bernoulli($\beta$)}
\end{equation}

where $\odot$ is the element-wise product, and $\beta$ is the dropout rate. Now, $\mathcal{M}^u_\theta$ is the neural network using a dropout and it produces the posterior probability $p(y | \mathbf{x} ; \mathbf{u}, \theta, T)$. For brevity, we denote $p^{\mathbf{u}}_\theta(y|\mathbf{x}) := p(y | \mathbf{x} ; \mathbf{u}, \theta, T)$. Similarly, we can also extract an additional feature vector $\mathbf{v} \odot f(\mathbf{x})$, where $v^{j}$ $\sim$ Bernoulli($\beta$). Thus, we can define $p^\mathbf{v}_\theta(y|\mathbf{x})$.

We propose a new regularization loss to distill knowledge by reducing the KL-divergence between two logits $\mathcal{M}^{\mathbf{u}}_\theta(\mathbf{x})$ and $\mathcal{M}_\theta^{\mathbf{v}}(\mathbf{x})$. Our method is visualized in Figure \ref{fig:self KD methods}. This method has computational advantages because, unlike conventional methods, it uses a single existing model, does not require additional modules, shares an encoder, and only requires post fully connected layer operations. Because the two features have no superior relationship with each other, we use this loss in a symmetric manner.

As a result, we use the forward and \red{reverse} KL-divergence of both instances. Further discussion on this matter is provided in Section \ref{subsec: inclusive kl div}.
Formally, given an input data $\mathbf{x}$, label $y$, and randomly dropped operations $\mathbf{u}, \mathbf{v}$, the loss of the SD-Dropout method is defined as follows:
\begin{equation}
\begin{aligned}
    \LL_{SDD}(\mathbf{x}; \mathbf{u}, \mathbf{v}, \theta, T) := {} & \kld(p^\mathbf{u}_\theta(y|\mathbf{x})||p^\mathbf{v}_\theta(y|\mathbf{x})) \\
    &+ \kld(p^\mathbf{v}_\theta(y|\mathbf{x})||p^\mathbf{u}_\theta(y|\mathbf{x}))
\end{aligned}
\end{equation}

Our method matches the predictions of different dropout features from a single network, whereas the conventional knowledge distillation method matches predictions from a teacher and a student network. Thus, the total loss $\LL_{Total}$ is defined as follows:
\begin{equation}
\begin{aligned}
    \LL_{Total}(\mathbf{x}, y ; \mathbf{u}, \mathbf{v}, \theta, T) = {} & \LL_{CE}(\mathbf{x}, y; \; \theta) \\
    &+ \lambda_{SDD} \cdot T^{2} \cdot \LL_{SDD}(\mathbf{x}; \mathbf{u}, \mathbf{v}, \theta, T)
    \label{eq:loss with SD-Dropout}
\end{aligned}
\end{equation}

where $\LL_{CE}$ is the cross-entropy loss and $\lambda_{SDD}$ is the weight hyperparameter of the SD-Dropout method.

\subsection{Collaboration with other method}

Our method can easily collaborate with various self-knowledge distillation methods because it has no additional module or training scheme constraints. In collaboration, the loss can be described as Eq. (\ref{eq:loss with other methods}), where $\LL_{KD}$ is an additional self-knowledge distillation loss for collaboration.
\begin{equation}
\begin{aligned}
    \LL_{Total}(\mathbf{x}, y; \mathbf{u}, \mathbf{v}, \theta, T) = {} &\LL_{CE}(\mathbf{x}, y; \; \theta) \\
    &+ \lambda_{SDD} \cdot T^{2} \cdot \LL_{SDD}(\mathbf{x}; \mathbf{u}, \mathbf{v}, \theta, T) \\
    &+ \lambda_{KD} \cdot T^{2} \cdot \LL_{KD}(\mathbf{x} ; \theta, T)
\end{aligned}
\label{eq:loss with other methods}
\end{equation}

where $\lambda_{KD}$ is the weight hyperparameter of the other distillation methods. The discussion of the appropriate $\lambda_{KD}$ value is detailed in Appendix \ref{subsection: hyper-parameter}.

\subsection{Forward versus \red{Reverse} KL-Divergence} \label{subsec: inclusive kl div}
Let $p_{\theta}(\mathbf{x})$ and $q_{\theta}(\mathbf{x})$ be the probability distributions. Let $N$ denote the size of input vector $\mathbf{x}$. Then, two kinds of KL-divergence, forward and \red{reverse} KL-divergence, are defined as follows:
\begin{align}
    D_{KL}^{fw.}(p_\theta,q_\theta) = D_{KL}(p || q_\theta) + D_{KL}(q || p_\theta) \\
    D_{KL}^{rv.}(p_\theta,q_\theta) = D_{KL}(p_\theta || q) + D_{KL}(q_\theta ||p ) 
\end{align}
Note that the absence of $\theta$ in $p$ indicates that $p$ is considered constant with respect to $\theta$, meaning that the gradient is not propagated. 

In the field of knowledge distillation, it is widely accepted that the forward KL-divergence is based on similarity with the Cross-Entropy loss. That is, the features of the teacher network are considered as the ground-truth, and the features of the student network are considered as logits that approximate the features of the teacher networks. However, we also adopt a \red{reverse} KL-divergence direction to further reduce the divergence between the two distributions. 

\red{The idea of utilizing reverse direction is also proposed as a new loss approximating the target Dirichlet distribution in \cite{malinin2019reverse_kl}. Since forward KL divergence  is zero-avoiding, utilizing only the forward direction is not suitable if the target Dirichlet distribution is multi-modal. We extend the analysis of the direction of KL divergence to the general distribution settings.}

\red{We claim that the derivative of reverse divergence is stronger than that of forward divergence. We show this by proving Proposition \ref{prop1} with analysis presented below. This indicates that the reverse direction is not negligible and plays an important role. In addition, we claim that the forward and reverse KL divergence work differently, in other words, their directions of the derivatives are quite different. We empirically verify this claim in Section \ref{subsection:dirKL}. Overall, by adding reverse divergence, we expect stronger self-knowledge distillation. }

First, we observe the representation of the derivatives of forward and \red{reverse} KL-divergence.
\begin{lemma}
The derivatives of forward and \red{reverse} divergence is represented as follows:
{\footnotesize 
\begin{equation}
    \nabla_{\theta}  D_{KL}^{fw.}(p_\theta,q_\theta)  = \sum_{i=1}^{N} (1 - \frac{p(\mathbf{x})_i}{q(\mathbf{x})_i} ) \nabla_\theta q(\mathbf{x})_i + \sum_{i=1}^{N} (1 - \frac{q(\mathbf{x})_i}{p(\mathbf{x})_i} ) \nabla_\theta p(\mathbf{x})_i 
\end{equation}
\begin{equation}
    \nabla_{\theta} D_{KL}^{rv.}(p_\theta,q_\theta)  = \sum_{i=1}^{N}  \log ( \frac{p(\mathbf{x})_i}{q(\mathbf{x})_i} )  \nabla_\theta p(\mathbf{x})_i + \sum_{i=1}^{N}  \log ( \frac{q(\mathbf{x})_i}{p(\mathbf{x})_i} )  \nabla_\theta q(\mathbf{x})_i
\end{equation}
}
\end{lemma}

Before beginning the main proposition, we make the following assumptions.

\begin{assumption} \label{ass1}
If $ | p(\mathbf{x})_i | > | q(\mathbf{x})_i | , $ then $ | \nabla_{\theta} p(\mathbf{x})_i | >  | \nabla_{\theta} q(\mathbf{x})_i | $. If $ | p(\mathbf{x})_i | < | q(\mathbf{x})_i | $, then $ | \nabla_{\theta} p(\mathbf{x})_i | < | \nabla_{\theta} q(\mathbf{x})_i | $.
$\textit{i.e.} ,  ( |\frac{p(\mathbf{x})_i}{q(\mathbf{x})_i}| - 1 ) (|\frac{ \nabla_{\theta} p(\mathbf{x})_i}{ \nabla_{\theta} q(\mathbf{x})_i}| - 1  ) > 0. $
\end{assumption}

\begin{assumption} \label{ass2}
Let $r = \log( | \frac{p(\mathbf{x})_i}{q(\mathbf{x})_i} | ) $ and  $ \rho = | \frac{\nabla_{\theta} p(\mathbf{x})_i}{\nabla_{\theta} q(\mathbf{x})_i} |> 1 $. Then $ r \le r_1$ where $r_1 = |\log(\rho) +\log(\log(\rho+(e-1))) | $.
\end{assumption}
Assumption \ref{ass1} implies that if $|p(\mathbf{x})|$ is greater than $| q(\mathbf{x})|$, then the derivative $ \nabla_{\theta} p(\mathbf{x}) $ is also greater than  $  \nabla_{\theta} q(\mathbf{x}) $. 
Assumption \ref{ass2} implies that the ratio  $| \frac{p(\mathbf{x})}{q(\mathbf{x})} |$ is not significantly different from the ratio $ | \frac{ \nabla_{\theta} p(\mathbf{x})}{\nabla_{\theta} q(\mathbf{x})} | $. 
We empirically validate that Assumptions \ref{ass1} and \ref{ass2} hold in probability during the experiment. 
Now, we posit the main proposition indicating that the \red{reverse} derivative is greater than the forward derivative under Assumptions \ref{ass1} and \ref{ass2}. 
In other words, we can demand a stronger connectedness (or bond) between logits $p(\mathbf{x})$ and $q(\mathbf{x})$ in the training process by adding \red{reverse} derivatives.
\begin{proposition} \label{prop1}
Under Assumptions \ref{ass1} and \ref{ass2}, let :
\begin{align}
\nonumber (D_{i,\theta}) &= |  \log( \frac{p(\mathbf{x})_i}{q(\mathbf{x})_i} )   \nabla_{\theta} q(\mathbf{x})_i  | + |  \log( \frac{q(\mathbf{x})_i}{p(\mathbf{x})_i} ) \nabla_{\theta} p(\mathbf{x})_i   |    \\
 - &  ( \; | (1 - \frac{p(\mathbf{x})_i}{q(\mathbf{x})_i} ) \nabla_{\theta} q(\mathbf{x})_i   | + | (1 - \frac{q(\mathbf{x})_i}{p(\mathbf{x})_i} ) \nabla_{\theta} p(\mathbf{x})_i   | \; )
\end{align}
Then we have:
\begin{equation}
    (D_{i,\theta}) > 0. 
\end{equation}
Moreover, $(D_i)$ has a maximum value at $r = |\log(\rho)|$. Here, $(D_i)$ implies the difference between the $L_1$ norm of the \red{reverse} derivatives and the $L_1$ norm of the forward derivatives. 

\end{proposition}
We present the detailed proofs of the above proposition in the supplementary materials.

\begin{table*}[t]
\caption{\red{Accuracy (\%) of self-knowledge distillation methods on CIFAR-100, CUB-200-2011, and Stanford Dogs datasets. Best results are indicated in bold.}}

\begin{center}
\begin{tabular}{|c|c|c|c|c|c|c|}
\hline

\multirow{2}{*}{Method} & \multicolumn{2}{c|}{CIFAR-100} & \multicolumn{2}{c|}{CUB-200-2011} & \multicolumn{2}{c|}{Standford Dogs} \\ \cline{2-7}
& Base & +SD-Dropout & Base & +SD-Dropout & Base & +SD-Dropout \\ \hline

Cross-Entropy & 74.8 & \textbf{77.0} (+2.2) & 53.8 & \textbf{66.6} (+12.8) & 63.8 & \textbf{69.9} (+6.1) \\
CS-KD & 77.3 & \textbf{77.4} (+0.1) & 64.9 & \textbf{65.4} (+0.6) & 68.8 & \textbf{69.3} (+0.5) \\
DDGSD & 76.8 & \textbf{77.1} (+0.3) & 58.3 & \textbf{62.9} (+4.6) & 66.9 & \textbf{68.1} (+1.3) \\
BYOT & 77.2 & \textbf{77.7} (+0.5) & 60.6 & \textbf{68.7} (+8.1) & 68.7 & \textbf{71.2} (+2.5) \\
DML & \textbf{78.9} & 78.8 (-0.1) & 61.5 & \textbf{65.7} (+4.2) & 70.5 & \textbf{72.0} (+1.6) \\
LS & 76.8 & \textbf{76.9} (+0.1) & 56.2 & \textbf{67.6} (+11.5) & 65.2 & \textbf{70.1} (+4.9) \\ \hline
\end{tabular}
\label{table:classification_result}
\end{center}
\end{table*}

\section{Experiments}\label{section:Experiments}
In this section, we present the effectiveness of the proposed method. \red{We demonstrate our method on a variety of tasks including image classification, object detection, calibration effect, robustness, and out-of-distribution detection. In addition, we conduct extensive experiments on the direction of KL-divergence, backbone networks, comparison with standard dropout, and generalizability of our method.}

\red{Throughout this section, we use ResNet-18 or its variants as the architecture and CIFAR-100 as the training dataset if there is no specific description. All experiments are performed using PyTorch \cite{NEURIPS2019_9015pytorch}.}

\subsection{Classification}

\subsubsection{CIFAR-100, CUB200, and Stanford Dogs} \label{subsection: cifar}

\paragraph{\textbf{Dataset}}
To validate the general performance of our method, we test various datasets including CIFAR-100 \cite{cifar}, CUB-200-2011 \cite{cub}, and Stanford Dogs \cite{stanforddog}.
CIFAR-100 is composed of 100 classes with large contextual differences between classes. By contrast, CUB-200-2011 and Stanford Dogs are composed of 200 and 120 fine-grained classes, and unlike CIFAR-100, there are smaller contextual differences between classes. 
The Performance in various dataset domains can be used to evaluate the overall performance of the model.

\paragraph{\textbf{Hyperparameters}}
For a fair comparison, we use the same hyperparameters in all experiments unless specifically mentioned. 
We use a stochastic gradient descent optimizer (learning rate=0.1, momentum=0.9, weight decay=1e-4) and train 200 epochs during all experiments. 
The learning rate is scheduled for decay 0.1 on 100 and 150 epochs. 
As a common setting for CIFAR-100, we set a batch size of 128 and use ResNet, which modifies the first convolution layer with a 3 $\times$ 3 kernel instead of a 7 $\times$ 7 kernel. 
We set input image sizes of 224 $\times$ 224 and batch sizes of 32 on CUB-200-2011 and Stanford Dogs.
All results are the average values obtained by repeating the experiment three times.

\paragraph{\textbf{Training Procedure}}
The training procedure of SD-Dropout is summarized as PyTorch-like style pseudo code, as described in Algorithm \ref{alg:code}. It should be noted that we do not use the .detach() method to calculate the KL-divergence terms to maintain both directions of KL-divergence.

\begin{algorithm}[t]
    \caption{\small{Pseudo code of SD-Dropout in a PyTorch-like style.}}
    \label{alg:code}
    \definecolor{codeblue}{rgb}{0.25,0.5,0.5}
    \lstset{
    	backgroundcolor=\color{white},
    	basicstyle=\fontsize{7.2pt}{7.2pt}\ttfamily\selectfont,
    	columns=fullflexible,
    	breaklines=true,
    	captionpos=b,
    	commentstyle=\fontsize{7.2pt}{7.2pt}\color{codeblue},
    	keywordstyle=\fontsize{7.2pt}{7.2pt},
    }
    \vskip -0.075in
    \begin{lstlisting}[language=python]
# x: input image
# y: ground-truth label
# backbone : feature extractor composed of convolutions
# classifier : fully connected layer
# lambda: weight hyperparamter of knowledge distillation method
for (x, y) in Batches:
    # extract the feature
    feat = backbone(x)
    # the values of the logits
    output = classifier(feat)
    # calculate cross-entropy loss
    loss_ce      = ce_loss(output, y)
    # sampling two features by dropout
    feat_dps = [dropout(feat) for _ in range(2)]
    # the logits of two dropout sampled feature
    output_dp1, output_dp2 = 
        [classifier(feat_dps[i]) for i in range(2)]
    # Foward reverse KL-divergence between two logits 
    loss_kd1  = kl_div_loss(output_dp1, output_dp2)
    loss_kd2  = kl_div_loss(output_dp2, output_dp1)
    loss_kd   = loss_kd1 + loss_kd2
    # total loss
    loss_total = loss_ce + lambda*loss_kd
    # update
    loss_total.backward()
    \end{lstlisting}
    \vskip -0.075in
\end{algorithm}

\paragraph{\textbf{Results}}

While using the fixed backbone model (ResNet-18), we compare the accuracy of the methods and datasets.
Table \textcolor{red}{\ref{table:classification_result}} shows the accuracy on CIFAR-100, CUB-200-2011, and Stanford Dogs in comparison with distillation and regularization methods, i.e., Cross-Entropy, CS-KD, DDGSD, BYOT, DML, and label smoothing (LS). 
Moreover, the [+ SD-Dropout] column is the result of a collaboration between existing methods. The dropout collaboration improves the performance of all datasets and self-distillation methodologies. 
Compared to previous self-distillation methods, although simply applicable, the SD-Dropout method performs best on the CUB-200-2011 dataset at 66.6\%. Furthermore, the largest increase in the performance showed that the compatibility with BYOT is the best (0.5\% on CIFAR-100, 8.1\% on CUB-200-2011, and 2.5\% on Stanford Dogs).

\begin{table}[t]
\caption{\red{Expected Calibration Error} comparison results of SD-Dropout combined with various knowledge distillation methodologies. \red{The trained network is ResNet-18. Lower is better and the best results are indicated in bold.}}

\begin{center}
\begin{tabular}{|c|c|c|}
\hline
Method & Base & +SD-Dropout\\
\hline
Cross-Entropy     & 0.120 & \textbf{0.075} \\
DDGSD  & 0.067 & \textbf{0.034} \\
CS-KD  & 0.068 & \textbf{0.046} \\
BYOT   & 0.117 & \textbf{0.056} \\
DML    & 0.058 & \textbf{0.039} \\
\hline
\end{tabular}
\label{table:ECE results}
\end{center}
\end{table}

\subsubsection{ImageNet}\label{sec:imagenet}

\begin{table}[t]
\caption{\red{Classification Accuracy (\%) on the ImageNet dataset for ResNet-152. Metric is Top-k accuracy (ACC @ k)}}
\begin{center}
\red{
\begin{tabular}{|c | c | c|}
\hline
Method & Acc @ 1 & Acc @ 5 \\
\hline
Cross-Entropy & 74.8 & 92.5 \\
SD-Dropout    & \textbf{75.5} & \textbf{92.7} \\
\hline
\end{tabular}
}
\label{tab: imagenet}
\end{center}
\end{table}

\paragraph{\textbf{Dataset}}

\red{
To validate our methodology even on the large-scale dataset, we evaluate the classification problem on ILSVRC 2012 dataset \cite{ImageNet}.
ImageNet is a dataset of 1000 classes of 1.28 million training images and 50k validation images.
We evaluate both top-1 and top-5 error rates.
}

\paragraph{\textbf{Hyperparameters}}
\red{
The hyperparameters used for learning baseline model and SD-dropout methodologies are as follows.
All models are trained for 90 epochs.
We use a 32 mini-batch size with a stochastic gradient descent optimizer.
The learning rate is scheduled for decay 0.1 on 30 and 60 epochs.
We use 1e-4 weight decay and 0.9 momentum.
The input image is randomly cropped with 224 $\times$ 224 size and augmented with random horizontal flip.
The lambda of the SD-Dropout method is 0.1.
All models used the same hyperparameters except for lambda.
The learning process is the same as section \ref{subsection: cifar}.
}

\paragraph{\textbf{Results}}

\red{
We select the ResNet-152 model as the baseline model to ensure that our method is valid even for the  large-scale dataset and large model.
Table \ref{tab: imagenet} shows that the accuracy of the SD-Dropout method is better than the result of learning the baseline model with cross-entropy loss.
The result implies that SD-Dropout is a universally applicable method for large datasets and large models as well as narrow datasets and small models.
}

\subsection{Object Detection}

\red{To verify the effectiveness of SD-Dropout on the visual recognition tasks, we perform the experiment on the task of object detection using the COCO dataset \cite{lin2014microsoft}. We use the Faster R-CNN \cite{ren2015faster} detector with a backbone of ResNet-152 trained on ImageNet as described in Section \ref{sec:imagenet}. We finetune Faster R-CNN for 7 epochs (0.5 $\times$ schedule) and 12 epochs (1 $\times$ schedule).
As shown in Table \ref{tab:detection}, SD-Dropout shows better results in object detection than the baseline. These results indicate that SD-Dropout is successful for transferring to another vision task.
}

\begin{table}[t]
\caption{\red{Results on object detection. mAP@0.5 denotes mean average precision with IOU threshold 0.5. mAP denotes COCO-style mAP. Best results are indicated in bold.} }
\begin{center}
\red{
\begin{tabular}{|c | c c|}
\hline
Method & 0.5 $\times$ Schedule & 1 $\times$ Schedule \\
\hline
& \multicolumn{2}{c|}{mAP / mAP@0.5} \\
Cross-Entropy & 31.2 / 50.6 & 39.4 / 60.1 \\
SD-Dropout & \textbf{32.5} / \textbf{52.8} & \textbf{39.8} / \textbf{60.7} \\
\hline
\end{tabular}
}
\label{tab:detection}
\end{center}
\end{table}

\subsection{Calibration Effect}
\begin{figure*}
     \centering
         \includegraphics[width=0.18\textwidth]{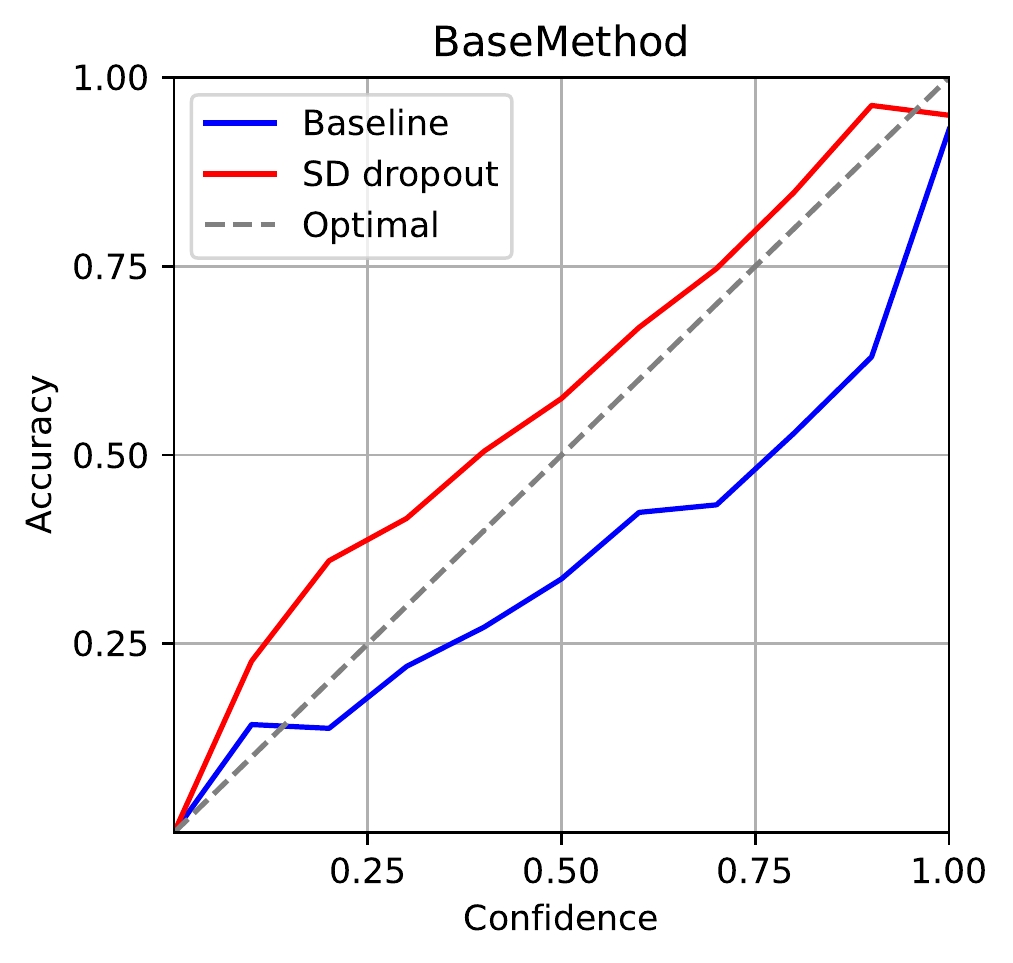}
         \includegraphics[width=0.18\textwidth]{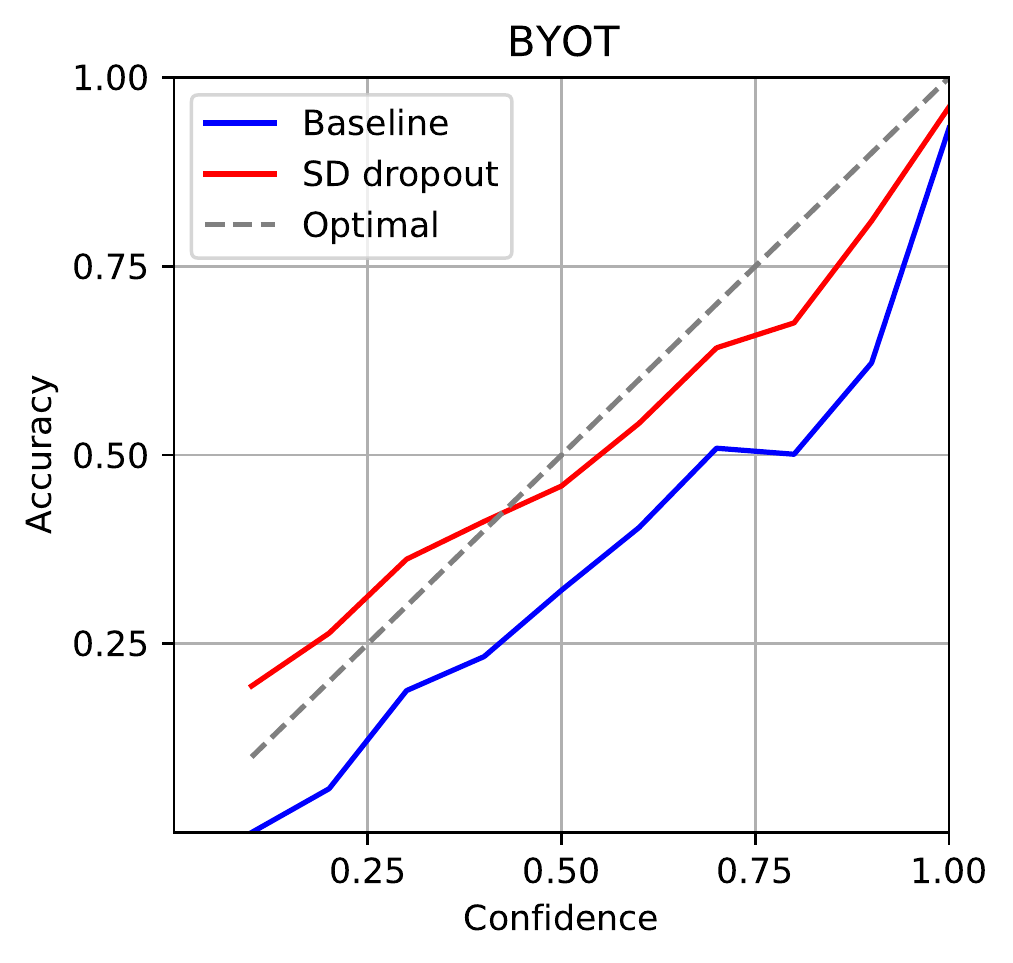}
         \includegraphics[width=0.18\textwidth]{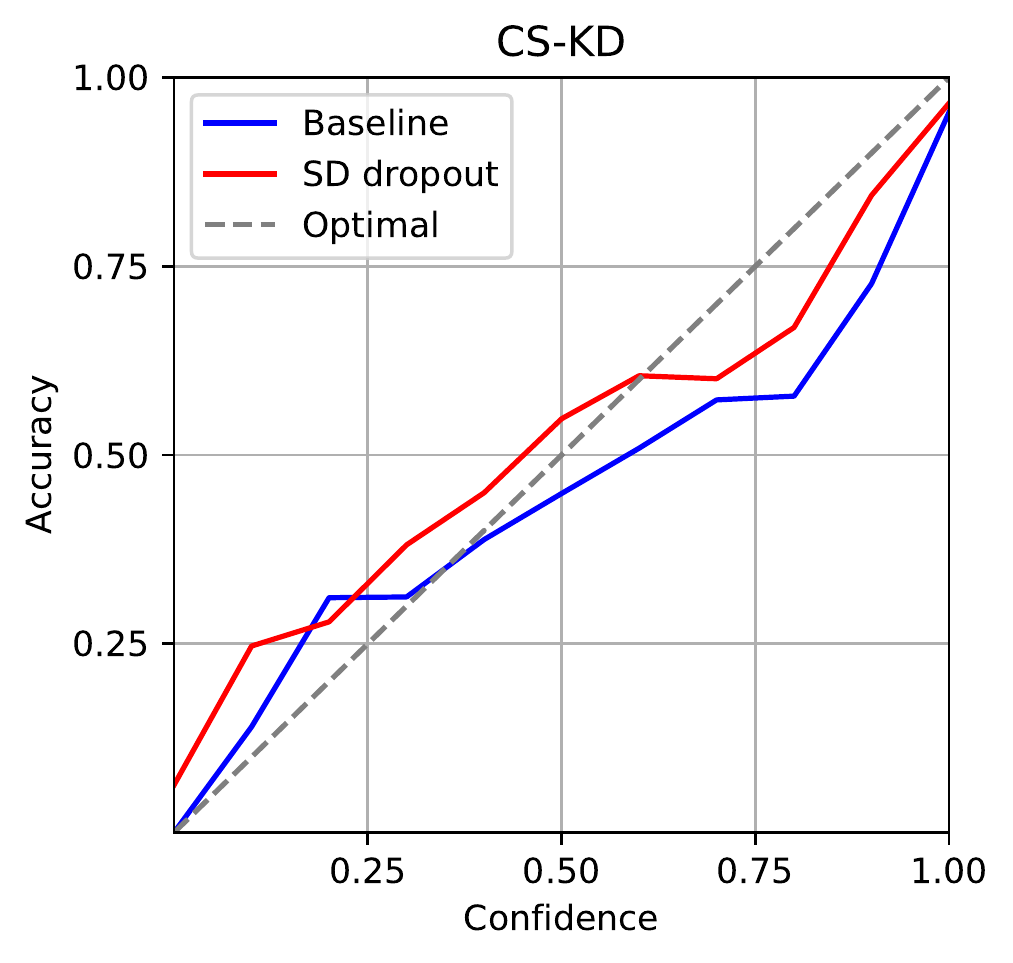}
         \includegraphics[width=0.18\textwidth]{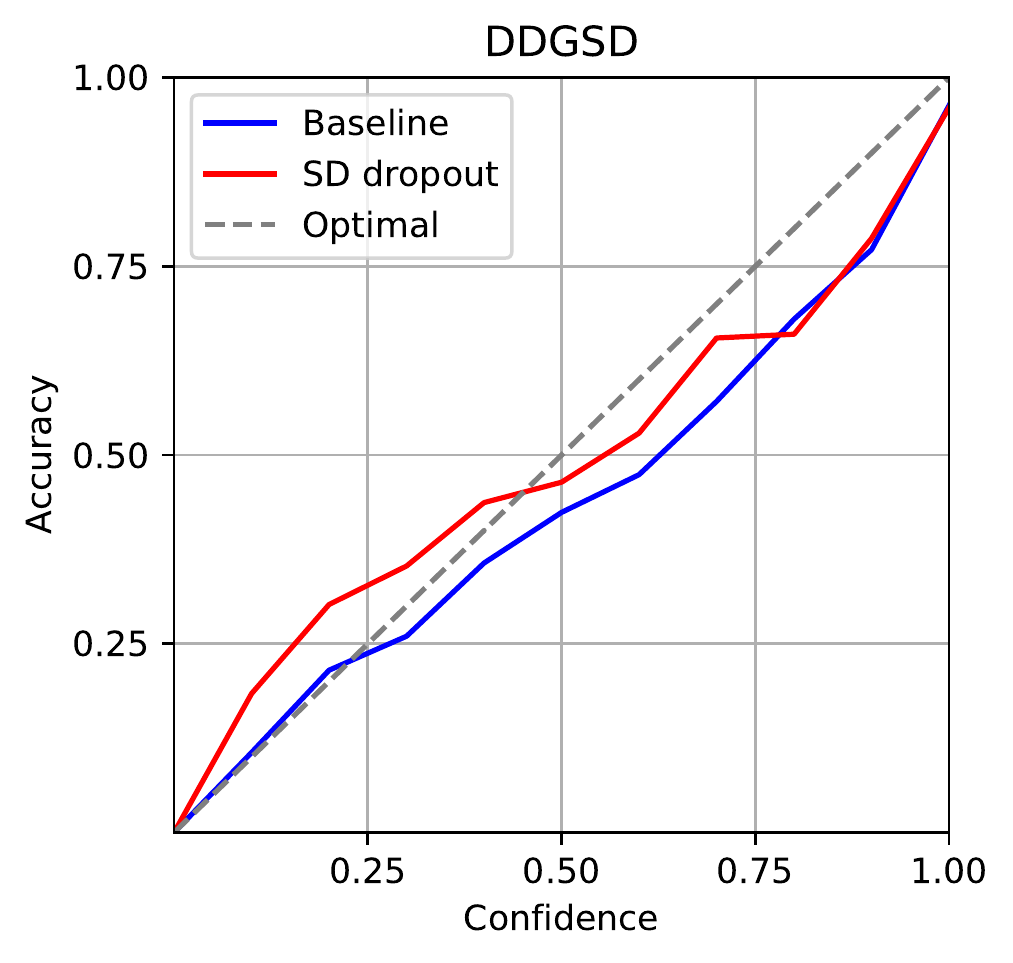}
         \includegraphics[width=0.18\textwidth]{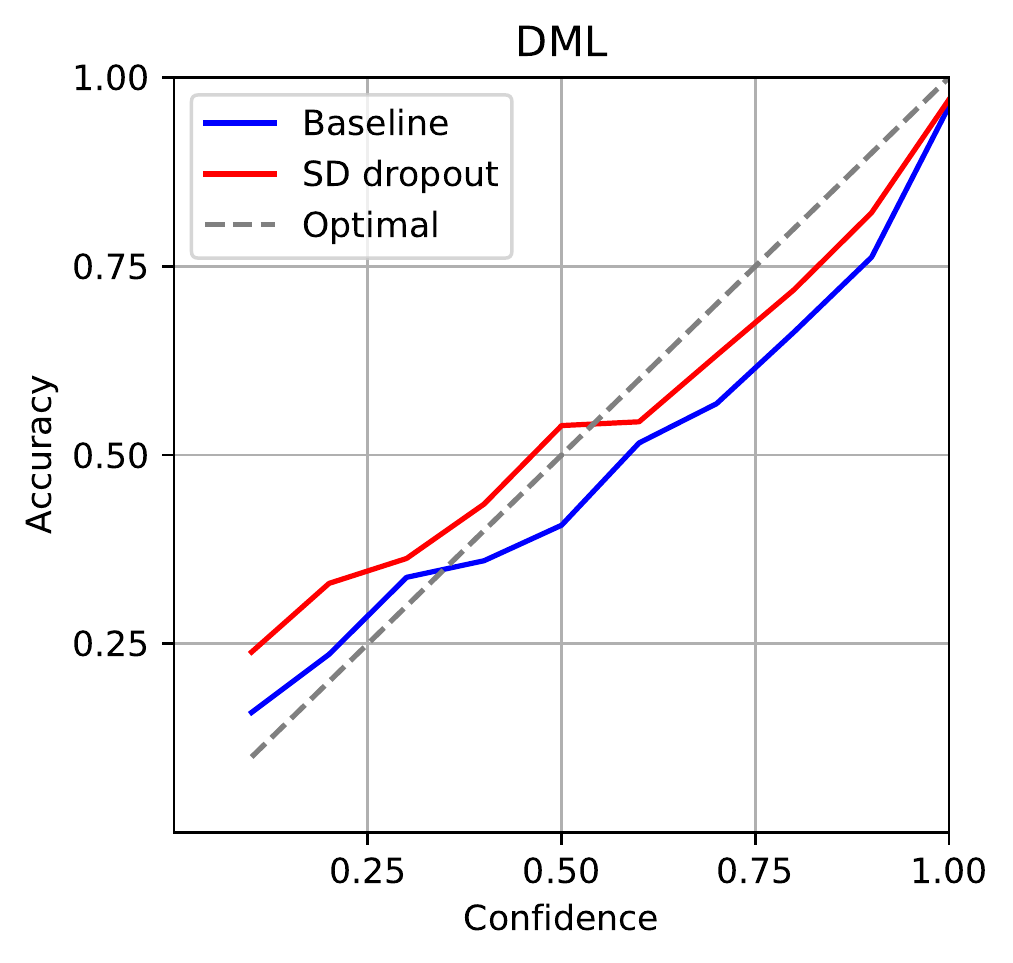}
    \caption{Reliability diagrams \cite{niculescu2005predicting-ecechart}. The x-axis is confidence bin and the y-axis is average accuracy on bin. The blue and red line show average accuracy w/ and w/o SD-dropout method.}
         
    \label{figure:ECE}
\end{figure*}

The expected calibration error (ECE) \cite{naeini2015obtaining-ece} is a metric that shows the difference between the confidence of the model predictions and the actual accuracy. The ECE can be calculated as
\begin{equation}
   ECE = \sum_{m=1}^{M} \frac{|B_m|}{n} \; |acc(B_m) - conf(B_m)|,  
\end{equation} 
where $M$, $n$, $B_m$, $acc(B_m)$, and $conf(B_m)$ denote the number of bins, the number of total samples, the number of samples in the $m$-th bin, the average accuracy of samples in the bin, and the average model confidence of samples in the bin.
We set the number of bins to 10. The results shown in Table \textcolor{red}{\ref{table:ECE results}} and Figure \textcolor{red}{\ref{figure:ECE}} indicate that SD-Dropout suppresses overconfidence and improves confidence calibration.

\subsection{Robustnetss}

\subsubsection{Robustness to adversarial attack}

\begin{table}[t]
\caption{Robustness comparison against the adversarial attack. Accuracy (\%) of ResNet-18 on various datasets. \red{ Best results are indicated in bold.}}
\begin{center}
\begin{tabular}{|c|c|c|}
\hline
Dataset & Base & +SD-Dropout\\
\hline
CIFAR-100 & 37.9 & \textbf{47.1}\\
CUB-200-2011	& 17.0 & \textbf{24.8}\\
Stanford Dogs & 19.2 & \textbf{22.6}\\
\hline
\end{tabular}
\label{table:adversarial robustness}
\end{center}
\end{table}

\begin{figure*}[t]
\begin{center}
   \includegraphics[width=0.9\textwidth,height=0.28\textwidth]{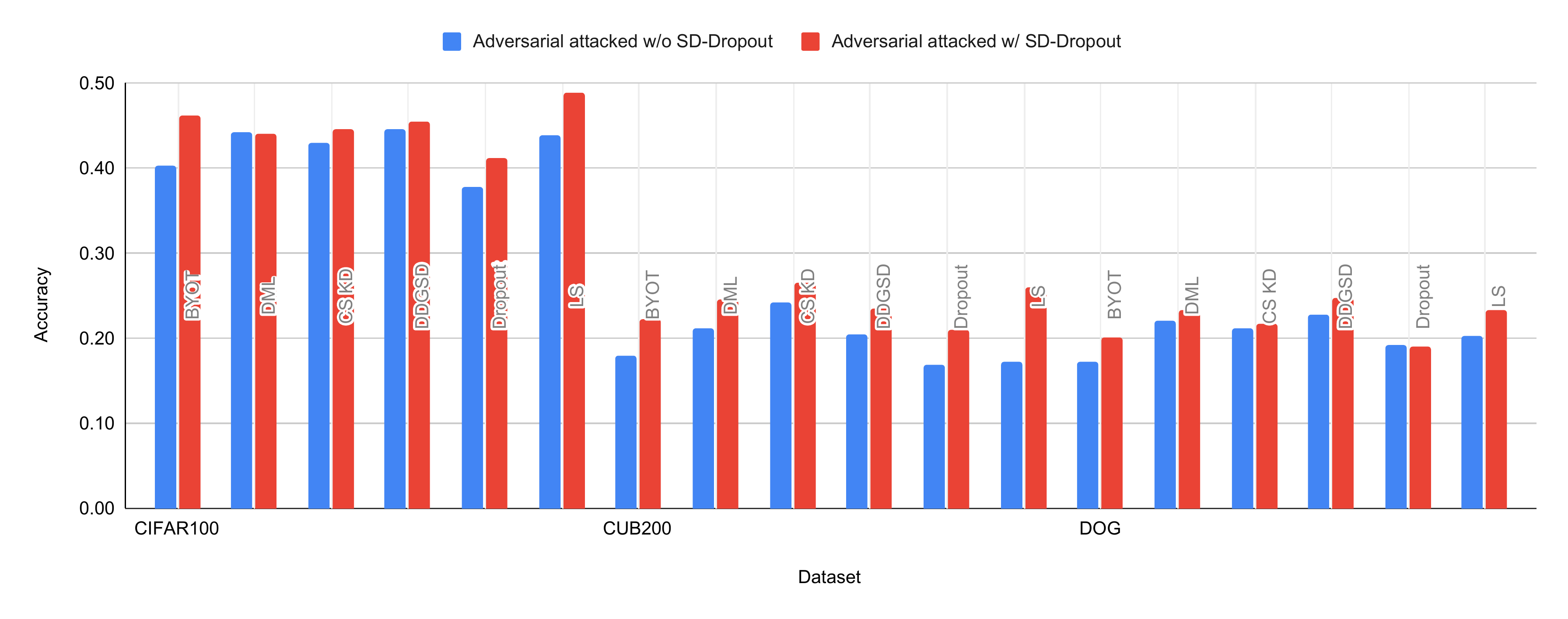}
\end{center}
   \caption{Adversarial robustness with collaborate cases. \red{The blue bar denotes the accuracy of base method without SD-dropout applied the adversarial attack. The red bar denotes the accuracy of collaborate method with SD-Dropout.}}
\label{fig:adversarial results chart}
\end{figure*}

To evaluate the robustness to adversarial attacks \cite{dong2020benchmarking-adversarial-benchmark}, 
\red{we compare the performance of methods subjected to an adversarial attack}.
An adversarial attack approach is the Fast gradient sign method \cite{goodfellow2014explaining-fgsm} that exploits the gradient of network with respect to the input image to increase the loss, where the maximum perturbation size is $\epsilon$ = 0.2. \red{The result are reported in Table \ref{table:adversarial robustness}.}

As shown in Figure \textcolor{red}{\ref{fig:adversarial results chart}}, we demonstrate the adversarial robustness with collaborate cases. 
All results show increases in the robustness of the adversarial attacks. It can be concluded that the SD-Dropout method has an effect similar to that of adversarial training.

\subsubsection{\red{Robustness to Distribution Shift}}

\begin{table}[t]
\caption{\red{Robustness to the distribution shift dataset, CIFAR-C. There are 15 types of corruption. mCA indicates the mean corruption accuracy. The network ResNet-18 is trained by the CIFAR-100 dataset. Best results are indicated in bold.}}
\begin{center}
\red{
\begin{tabular}{ |c|c|c|c| }
\hline
\multicolumn{2}{|c|}{Corruption} & Cross-Entropy & SD-Dropout \\
\hline
\multirow{3}{*}{Noise} & Gauss. & \textbf{23.1} & 21.9 \\
                       & shot & \textbf{31.2} & 30.2 \\
                       & Impulse & \textbf{27.1} & 25.4 \\
\hline
\multirow{4}{*}{Blur}  & Defocus & 57.0 & \textbf{59.1} \\
                       & Glass & \textbf{25.1} & 23.1 \\
                       & Motion & 52.4 & \textbf{54.2} \\
                       & Zoom & 50.1 & \textbf{52.9} \\
\hline
\multirow{4}{*}{Weather} & Snow & 52.7 & \textbf{54.9} \\
                       & Frost & 47.2 & \textbf{48.7} \\
                       & Fog & 61.1 & \textbf{63.3} \\
                       & Bright & 70.2 & \textbf{72.4} \\
\hline
\multirow{4}{*}{Digital} & Contrast & 50.6 & \textbf{53.6} \\
                       & Elastic & 59.4 & \textbf{60.7} \\
                       & Pixel & \textbf{52.9} & 52.7 \\
                       & JPEG & \textbf{52.4} & 51.2 \\
\hline
\multicolumn{2}{|c|}{\multirow{2}{*}{\textbf{mCA}}} & \multirow{2}{*}{47.49} & \multirow{2}{*}{\textbf{48.28}} \\
\multicolumn{2}{|c|}{} & & \\
\hline
\end{tabular}
}
\end{center}
\label{tab:cifar-c}
\end{table}

\red{To evaluate the classification robustness of the distribution shift task, we compare the accuracy of methods on the CIFAR-C dataset \cite{hendrycks2019benchmarking}. CIFAR-C dataset includes 15 types of corruption from noise, blur, weather, and digital categories. We evaluate the networks trained by the normal CIFAR-100 dataset. In table \ref{tab:cifar-c}, the average mean corruption accuracy (mCA) of our method outperforms the baseline. 
In detail, SD-Dropout is better for nine types of corruption, but worse for six types of corruption.
As can be seen from the experimental results, SD-Dropout is generally robust in various noise environments of the input images.}

\subsection{\red{Out-of-Distribution Task}} \label{sec: result of ood task}

\begin{table}[t]
\caption{\red{Result on the out-of-distribution dataset. The CIFAR-100 dataset is used as an in-distribution dataset. $\uparrow$ means larger is better, and $\downarrow$ means lower is better. The metrics are explained in Section \ref{sec: result of ood task}. The trained network is ResNet-18. Best results are indicated in bold.}}
\begin{center}
\red{
\resizebox{\columnwidth}{!}{
\begin{tabular}{ |c|c c c c c |}
\hline
\multirow{3}{*}{Dataset} & FPR  & Detection & \multirow{2}{*}{AUROC} & AUPR & AUPR \\
& (at 95\% TPR) & Error & & (in) & (out) \\
& $\downarrow$ & $\downarrow$ & $\uparrow$ & $\uparrow$ & $\uparrow$ \\
\hline
& \multicolumn{5}{c|}{\textbf{Cross-Entropy/SD-Dropout} (\%)} \\
LSUN & 74.7/\textbf{63.5} & 25.0/\textbf{22.9} & 82.0/\textbf{84.9} & 84.5/\textbf{84.8} & 78.5/\textbf{83.4} \\
iSUN & 76.0/\textbf{63.8} & 25.3/\textbf{23.0} & 82.0/\textbf{84.8} & \textbf{85.9}/85.6 & 76.0/\textbf{81.9} \\
DTD  & 82.7/\textbf{77.5} & 28.6/\textbf{27.4} & \textbf{77.3}/77.2 & \textbf{86.3}/83.7 & 59.8/\textbf{64.0} \\
SVHN & 82.6/\textbf{75.2} & \textbf{22.7}/27.4 & \textbf{82.9}/79.1 & \textbf{86.6}/78.3 & 75.5/\textbf{76.7} \\
\hline
\textbf{Average} & 79.0/\textbf{70.0} & 25.4/\textbf{25.1} & 81.1/\textbf{81.5} & \textbf{85.8}/83.1 & 72.4/\textbf{76.5} \\
\hline
\end{tabular}
}
}
\end{center}
\label{tab:ood}
\end{table}

\red{We also validate that our method can detect whether it is an out-of-distribution example. It is important whether our model can predict whether the test examples come from an in-distribution or an out-distribution. To evaluate the performance of out-of-distribution detection, we use the ODIN detector \cite{liang2017enhancing_ood}. The ODIN detector can discriminate in-distribution and out-distribution by using temperature scaling and adding small perturbations to a test example. Since the ODIN method does not require any changes to a pre-trained network, we use the baseline and SD-Dropout model as the backbone of the ODIN detector. The other hyper-parameters are set to the same values as the original paper.}

\red{We use four following metrics to evaluate the detection performance on in- and out-of-distribution datasets. \textbf{FPR at 95 \% TPR}) is the false positive rate (FPR) when the true positive rate (TPR) is 95\%. \textbf{Detection Error} is calculated by $0.5 \cdot (1-TPR) + 0.5 \cdot FPR$. \textbf{AUROC} is the area under the receiver operating characteristic curve. \textbf{AUPR} is the area under the precision-recall curve. We measure AUPR both in- and out-of-distribution dataset, respectively.
Here, the in-distribution dataset is the CIFAR-100 datasets. We use LSUN \cite{yu2015lsun}, iSUN \cite{xu2015isun}, DTD \cite{cimpoi2014dtd_dataset}, and SVHN \cite{netzer2011svhn} datasets as out-of-distribution datasets.}

\red{The result is shown in Table \ref{tab:ood}. 
SD-Dropout induces the reduction in the variation of the features sampled by the dropout. As a result, the proposed method reduces the uncertainty of in-distribution dataset.
This is similar to a contrastive loss of positive pairs in constrative learning.
For this reason, it can be concluded that our method shows superior performance in the out-of-distribution task compared to the cross-entropy.}

\subsection{Additional Study}

\subsubsection{Directions of KL-divergence} \label{subsection:dirKL}
We empirically verify through experiments that Assumptions \ref{ass1} and \ref{ass2} in Section \ref{subsec: inclusive kl div} are convincing (see Table \ref{table:assmpt}). We use ResNet-18 on the CIFAR-100 dataset in the experiment. The probability that Assumption \ref{ass1} holds is greater than 0.5 in all epoch. $L_1$ norm of $r$ in Assumption \ref{ass2} is smaller than $r_1$ in all epochs.

\begin{figure}
     \centering
         \includegraphics[width=0.4\textwidth]{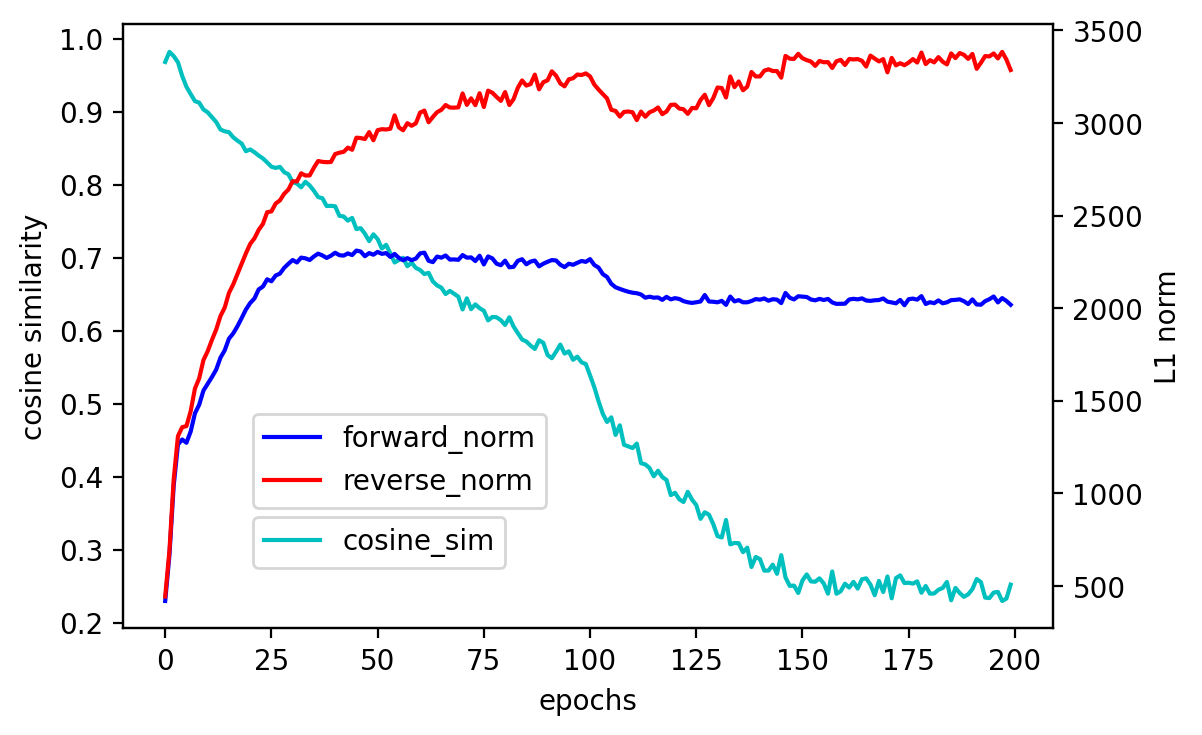}
    \caption{\red{Cosine Similarity between the gradients of forward and reverse KL divergence and $L_1$ norms.} }
    \label{figure:cossim}
\end{figure}

\red{As discussed in Section \ref{subsec: inclusive kl div}, we use both directions of KL divergence. To verify its effectiveness, we conduct an ablation experiment on the directions of KL divergence. As shown in Table \ref{table:directionkl},  the model using both directions of KL-divergence achieves the best performance. In addition, we investigate the directions of the gradient from forward and reverse KL divergence on the CIFAR-100. As shown in Figure \ref{figure:cossim}, the cosine similarity between the gradients in the two directions gradually decreases as training progresses. This indicates that the directions of two gradients are significantly different, and both directions of KL divergence are essential for training. Furthermore, we also see that the $L_1$ norm of the gradient from the reverse derivative is greater than the forward derivative, similar to the analysis in Section \ref{subsec: inclusive kl div}.  }

\begin{table}[t]
    \caption{Verification of assumptions in Section \ref{subsec: inclusive kl div}}
\begin{center}
\begin{tabular}{|c|c|c|c|}
\hline
Epoch    & $P$(Assumption \ref{ass1}) &  $r$  & $r_1$    \\ 
\hline
0         & 0.638 & 0.0681 & 0.1368  \\
100       & 0.653 & 0.1403 & 0.2799  \\
200       & 0.594 & 0.1292 & 0.2662  \\  
\hline
\end{tabular}
\end{center}
\label{table:assmpt}
\end{table}

\begin{table}[t]
    \caption{\red{Accuracy (\%) comparison between the forward, reverse, and both directions (forward and reverse) of KL-divergence on ResNet-18. Best results are indicated in bold.}}
\begin{center}
\red{
\resizebox{\columnwidth}{!}{
\begin{tabular}{|c|c|c|c|c|}
\hline
Dataset & Base & Forward & Reverse & Both Directions  \\
\hline
CIFAR-100 & 74.8 & 76.6  & 76.3 & \textbf{77.0} \\
CUB-200-2011 & 53.8 & 65.4& 63.8  & \textbf{66.6}  \\
Stanford Dogs & 64.1 & 69.6& 69.7 & \textbf{69.8} \\  
\hline
\end{tabular}
}
}
\end{center}
\label{table:directionkl}
\end{table}

\subsubsection{Backbone Network}

\begin{table}[t]
\caption{Accuracy (\%) comparison with different backbone networks on CIFAR-100. \red{Best results are indicated in bold.}}
\begin{center}
\begin{tabular}{|c|c|c|}
\hline
Architecture & Base & +SD-Dropout \\
\hline
ResNet-18 & 74.8 & \textbf{77.0} \\
ResNet-34 & 75.7 & \textbf{77.2}  \\
DenseNet-121 & 77.3 & \textbf{78.4}  \\
\hline
\end{tabular}
\label{table:model variant}
\end{center}
\end{table}

Our self-distillation method is easily adaptable to various backbone models. We compare several backbone networks with and without SD-Dropout. We apply SD-Dropout to ResNet-18, ResNet-34, and DenseNet-121. Table \ref{table:model variant} shows that the SD-Dropout method can improve the network performance regardless of the backbone networks. In particular, our SD-Dropout improves the accuracy of the baseline networks from 74.8\% to 77.0\% for ResNet-18, and from 75.7\% to 77.2\% for ResNet-34 on the CIFAR-100 dataset. For DenseNet-121, \red{our method enhances the accuracy from 77.3\% to 78.4\%}.

\begin{table}[t]
\caption{\red{Accuracy (\%) comparison between standard dropout and SD-Dropout. [standard$^\dagger$] represents the standard dropout method that equalizes the number of training steps compared to SD-Dropout. Best results are indicated in bold.}}
\begin{center}
\red{
\resizebox{\columnwidth}{!}{
\begin{tabular}{|c|c|c|c|c|}
\hline
\multirow{2}{*}{Dataset} & \multirow{2}{*}{Base} & +Dropout & +Dropout & \multirow{2}{*}{+SD-Dropout} \\
& & (standard) & (standard$^\dagger$) & \\
\hline
CIFAR-100 & 74.8 & 75.4 & 73.5 & \textbf{77.0} \\
CUB-200-2011 & 53.8 & 64.6 & 60.3 & \textbf{66.6}  \\
Stanford Dogs & 64.1 & 69.5 & 68.0 & \textbf{69.8}  \\
\hline
\end{tabular}
}
}
\label{table:comp_standard_dropout}
\end{center}
\end{table}

\begin{table}[t]
\caption{\red{Accuracy comparison with various positions of dropout on CIFAR-100. The trained network is ResNet-18. Best results are indicated in bold.}}
\begin{center}
\red{
\begin{tabular}{|c|c|}
\hline
Method & Accuracy (\%) \\
\hline
Cross-Entropy & 74.8  \\
SD-Dropout & \textbf{77.0}   \\
SD-Dropout (after layer1) & 74.8 \\
SD-Dropout (after layer2) & 75.1 \\
SD-Dropout (after layer3) & 75.7 \\
\hline
\end{tabular}
}
\label{tab:different layers}
\end{center}
\end{table}

\begin{table}[t]
\caption{\red{Accuracy comparison between networks with multiple fully-connected layers on CIFAR-100. The trained network is ResNet-18 with two fully-connected layers. Best results are indicated in bold.}}
\begin{center}

\red{
\begin{tabular}{|c|c|}
\hline
Method & Accuracy (\%) \\
\hline
Cross-Entropy & 74.9  \\
SD-Dropout    & \textbf{75.9}  \\
\hline
\end{tabular}}
\label{tab: multiple fc layers}
\end{center}
\end{table}

\subsubsection{Comparison \red{with Standard} Dropout}

The dropout technique plays the most important role in the SD-Dropout method. 
\red{To demonstrate the importance of dropout distillation with our method, we compare it with networks using standard dropout methods. Also, we equalize the number of training steps between standard dropout and SD-Dropout methods.} Table \ref{table:comp_standard_dropout} compares the experimental results of the \red{standard} dropout and our SD-Dropout methods on the CIFAR-100, CUB-200-2011, and Stanford Dogs datasets. It is observed that our SD-Dropout method outperforms the \red{
standard} dropout \red{methods} on all datasets. 

\subsubsection{\red{Dropout at Various Positions}}

\red{To get various sampled feature vectors, we apply dropout from deep layer to shallow layer. 
It is important which layer to apply the dropout. 
While our method applies dropout prior to a fully-connected layer, our method can be generalized and apply dropout to any layer, theoretically. 
We conduct further experiments from this perspective. 
In particular, ResNet can be divided into 4 large parts. 
For convenience, name these parts layer1$\sim$4. 
That is, our method applies dropout after layer4. 
In Table \ref{tab:different layers}, we compare the results of methods that apply dropout to a different layer. 
Although dropouts are applied to different layers, they mostly outperform the baseline cross-entropy method.
From the experimental results, the high-level features sampled from the deep layer are more suitable for SD-Dropout than the low-level features sampled from the shallow layer.}

\subsubsection{\red{Multiple Fully-connected Layers}}

\red{Our method is model-agnostic, which means it can be applied to any structure of the model. Throughout the paper, we treat the network whose structure is composed of convolutional layers and one fully-connected layer. To show that the SD-dropout method can be applied to any structure, we apply our method to a network with multiple fully-connected layers. We append one additional fully-connected layer with 512 hidden units. The result is in Table \ref{tab: multiple fc layers}. It appears that our method also improves the performance of a network with multiple fully-connected layers.}

\section{Conclusion}

We propose a new and simple self-knowledge distillation method.
The proposed method samples different models through a dropout and distills the knowledge of both.
We also experimentally and analytically show the characteristics of \red{reverse} KL-divergence.
We demonstrate that the proposed method improves generalization, calibration performance, adversarial robustness, \red{and ability of out-of-distribution detection.}
From the perspective of the regularization domain, our method is superior to the conventional label smoothing method through multiple datasets.
Thus, we expect our method to be used as a regularization method that can effectively improve the performance of a single network in various domains.

{\small
\bibliographystyle{ieee_fullname}
\bibliography{main}
}

\newpage
\appendix

\section{Appendix}

\subsection{Hyperparameters} \label{subsection: hyper-parameter}

Using SD-Dropout, we conduct additional experiments on ResNet-18 for the CIFAR-100 dataset to obtain the appropriate hyperparameters, where $\beta$ is the dropout rate and $\lambda_{SDD}$ is the weight of SD-Dropout in the loss function. We investigate $\beta \in \{0.1, 0.3, 0.5, 0.7 \}$ and $\lambda_{SDD} \in \{0.1, 0.5, 1.0, 2.0, 5.0\}$. We maintain the other conditions except for the hyperparameters $\beta$ and $\lambda_{SDD}$. Thus, we found the suitable hyperparameters $\beta = 0.5$ and $\lambda_{SDD} = 1.0$ in Table \ref{table:Hyaperpara}.

\subsection{Proof of Theorems}

\begin{statement}{Lemma 1}
The derivatives of forward and backward divergence are represented as follows:
{\footnotesize 
\begin{equation}
    \nabla_{\theta}  D_{KL}^{fw.}(p_\theta,q_\theta)  = \sum_{i=1}^{N} (1 - \frac{p(\mathbf{x})_i}{q(\mathbf{x})_i} ) \nabla_\theta q(\mathbf{x})_i + \sum_{i=1}^{N} (1 - \frac{q(\mathbf{x})_i}{p(\mathbf{x})_i} ) \nabla_\theta p(\mathbf{x})_i 
\end{equation}
\begin{equation}
    \nabla_{\theta} D_{KL}^{bw.}(p_\theta,q_\theta)  = \sum_{i=1}^{N}  \log ( \frac{p(\mathbf{x})_i}{q(\mathbf{x})_i} )  \nabla_\theta p(\mathbf{x})_i + \sum_{i=1}^{N}  \log ( \frac{q(\mathbf{x})_i}{p(\mathbf{x})_i} )  \nabla_\theta q(\mathbf{x})_i
\end{equation}
}
\end{statement}
(Proof)
Let $Z_q$ $=$ $\sum_{i=1}^{N} q(\mathbf{x})_i.$ Because $\sum_{i=1}^{N}$ $q(\mathbf{x})_i$ $= 1$, we have $Z_q=1$ and $q_\theta$ $= \frac{q_\theta}{Z_q}$. Then, we can calculate the forward derivative as
{\footnotesize 
    \begin{align}
   &  \nabla_{\theta} D_{KL}(p,q_\theta) \nonumber \\
   &= \nabla_\theta ( \sum_{i=1}^{N} p(\mathbf{x})_i ( \log p(\mathbf{x})_i - \log q_\theta(\mathbf{x})_i + \log Z_q ) ) \nonumber \\
   &= \sum_{i=1}^{N} p(\mathbf{x})_i ( -\frac{\nabla_\theta q_\theta(\mathbf{x})_i}{q_\theta(\mathbf{x})_i} + \frac{\nabla_\theta Z_q}{Z_q} ) \nonumber \\
   &= \sum_{i=1}^{N} -\frac{p(\mathbf{x})_i}{q(\mathbf{x})_i} \nabla_\theta q_\theta(\mathbf{x})_i +  \frac{\nabla_\theta Z_q}{Z_q} \sum_{i=1}^{N} p(\mathbf{x})_i
\end{align}
}
Because $Z_q = 1, \sum_{i=1}^{N} p(\mathbf{x})_i=1$, and $ \nabla_\theta Z_q =  \sum_{i=1}^{N} \nabla_\theta q(\mathbf{x})_i$, we finally obtain
\begin{equation}
     \nabla_{\theta} D_{KL}(p,q_\theta) =  \sum_{i=1}^{N} \nabla_\theta q(\mathbf{x})_i ( -\frac{p(\mathbf{x})_i}{q_\theta(\mathbf{x})_i} + 1) . 
\end{equation}
Similarly, we can calculate the following backward derivative: 
\begin{align}
& \nabla_{\theta} D_{KL}(p_\theta,q) \nonumber \\
& = \nabla_{\theta} \sum_{i=1}^{N} p_\theta (\mathbf{x})_i  (\log p_\theta (\mathbf{x})_i - \log Z_p - \log q (\mathbf{x})_i ) \nonumber \\
& = \sum_{i=1}^{N} \nabla_{\theta} p_\theta (\mathbf{x})_i (\log p_\theta (\mathbf{x})_i - \log Z_p - \log q(\mathbf{x})_i ) \nonumber \\
& + \sum_{i=1}^{N} p_\theta (\mathbf{x})_i (\frac{\nabla_{\theta} p_\theta (\mathbf{x})_i }{p_\theta (\mathbf{x})_i} - \frac{\nabla_{\theta} Z_p}{Z_p} ) \nonumber \\
& = \sum_{i=1}^{N}  \nabla_{\theta} p_\theta (\mathbf{x})_i (\log \frac{p_\theta (\mathbf{x})_i }{q (\mathbf{x})_i} - \log Z_p ) \nonumber \\
& + \sum_{i=1}^{N} \nabla_{\theta} p_\theta (\mathbf{x})_i  -  \frac{\nabla_{\theta} Z_p}{Z_p} \sum_{i=1}^{N} p_\theta (\mathbf{x})_i \nonumber \\
&= \sum_{i=1}^{N} \nabla_{\theta} p_\theta (\mathbf{x})_i \log \frac{p_\theta (\mathbf{x})_i }{q(\mathbf{x})_i}. \qed
\end{align}

\begin{statement}{Proposition 1}
Under Assumptions \ref{ass1} and \ref{ass2}, let
\begin{align}
\nonumber (D_i) &= |  \log( \frac{p(\mathbf{x})_i}{q(\mathbf{x})_i} ) \nabla_{\theta} q(\mathbf{x})_i  | + |  \log( \frac{q(\mathbf{x})_i}{p(\mathbf{x})_i} ) \nabla_{\theta} p(\mathbf{x})_i   |    \\
 - &  ( \; | (1 - \frac{p(\mathbf{x})_i}{q(\mathbf{x})_i} ) \nabla_{\theta} q(\mathbf{x})_i   | + | (1 - \frac{q(\mathbf{x})_i}{p(\mathbf{x})_i} ) \nabla_{\theta} p(\mathbf{x})_i   | \; )
\end{align}
Then, we have the following:
\begin{equation}
    (D_i) > 0. 
\end{equation}
Moreover, $(D_i)$ has the maximum value at $r = |\log(\rho)|$.  
Therefore if we take 
\begin{align}
\nonumber (D) &=  \|  \log( \frac{p(\mathbf{x})}{q(\mathbf{x})} ) \nabla_{\theta} q(\mathbf{x})  \| + \|  \log( \frac{q(\mathbf{x})}{p(\mathbf{x})} ) \nabla_{\theta} p(\mathbf{x})   \|_1  \\
 - & ( \;  \| (1 - \frac{p(\mathbf{x})}{q(\mathbf{x})} ) \nabla_{\theta} q(\mathbf{x})   \| + \| (1 - \frac{q(\mathbf{x})}{p(\mathbf{x})} ) \nabla_{\theta} p(\mathbf{x})   \|_1 \; )
\end{align}
We then have
\begin{equation}
     (D) > 0,
\end{equation}
where $(D)$ implies the difference between the $L_1$ norm of the backward derivatives and the $L_1$ norm of the forward derivatives.

\end{statement}

\begin{table}
\caption{Accuracy (\%) comparison of ResNet-18 on CIFAR-100 dataset over various hyperparameters $\beta$ and $\lambda_{SDD}$.}
\begin{center}
\begin{tabular}{|c|c|c|c|c|}
\hline
\backslashbox{$\lambda_{SDD}$}{$\beta$}& 0.1 & 0.3 & 0.5 & 0.7 \\
\hline
0.1 & 76.31 & 76.47 & 75.58 & 76.91 \\
0.5 & 75.83 & 76.31 & 76.88 & 76.43  \\
1.0 & 75.72 & 76.75 & \bf{77.10} & 76.82  \\
2.0 & 76.86 & 76.79 & 77.07 & 76.91 \\
5.0 & 76.92 & 76.79 & 76.57 & 69.47 \\
\hline
\end{tabular}
\label{table:Hyaperpara}
\end{center}
\end{table}

(Proof) 
For $i \in [N]$, without a loss of generality, we set $ |p(\mathbf{x})_i| \ge | q(\mathbf{x})_i| , \textit{that is}, r \ge 1$.  By Assumption \ref{ass1}, we take $  | \nabla_{\theta} p(\mathbf{x})_i| = \rho | \nabla_{\theta} q(\mathbf{x})_i| $ for $\rho >1$. Then,
\begin{align}
  &  | ( \log( \frac{p(\mathbf{x})_i}{q(\mathbf{x})_i} ) \nabla_{\theta} q(\mathbf{x})_i  | + | ( \log( \frac{q(\mathbf{x})_i}{p(\mathbf{x})_i} ) \nabla_{\theta} p(\mathbf{x})_i   | - \nonumber \\
  &  | (1 - \frac{p(\mathbf{x})_i}{q(\mathbf{x})_i} ) \nabla_{\theta} q(\mathbf{x})_i  | + | (1 - \frac{q(\mathbf{x})_i}{p(\mathbf{x})_i} ) \nabla_{\theta} p(\mathbf{x})_i   | = \nonumber \\
  &  [  ( 1 + \rho)r - \{ (e^r-1) + \rho ( 1- e^{-r}) \}   ] | \nabla_{\theta} q(\mathbf{x})_i|.
\end{align} Let 
\begin{equation}
   k(r) := ( 1 + \rho)r - \{ (e^r-1) + \rho ( 1- e^{-r}) \} 
\end{equation}
Then, we have $k'(r) = (1-e^r) + \rho (1-e^{-r})  = -e^{-r}( e^{2r} - (1+\rho)e^r + \rho) = -e^{-r} (e^r-1)(e^r-\rho)$. Thus, $k'(r)$ has roots at $r=0$ and $r=\log(\rho)$. Because $k(0)=k'(0)=0$, $k(r)$ increases in $r \in (0,\log(\rho)).$ Furthermore, $k(r)$ has a maximum value $k(\log(\rho))=(1+\rho)\log(\rho) -2\rho +2$ at $r= \log(\rho)$. \\
Now, we show $k(\log(\rho)+\log(\log(\rho+ (e - 1)  )) ) \ge 0$. Let $l(\rho) = k(\log(\rho)+\log(\log(\rho+ (e- 1 )  )) ) $, and  
\begin{align}
   \nonumber & l(\rho) = (\rho + 1 )(\log(\rho) + \log(\log(\rho+(e-1)))) \\
              &- \rho(\log(\rho+(e-1)) +1) + \log(\rho+(e-1))^{-1} +1.
\end{align}
The derivative of $l(\rho)$ is :
\begin{align}
     & l'(\rho) = -  ( (\rho+ (e-1))\log(\rho+(e-1))^2 )^{-1} \nonumber \\ 
     &  - ( \rho \log(\rho+(e-1)) )^{-1} - \rho(\rho+(e-1))^{-1} \nonumber \\ 
     &  + (\rho+1)(\rho+(e-1))^{-1}\log(\rho+(e-1))^{-1}  + 1 \nonumber \\
     & + \rho^{-1} -\log(\rho+(e-1))  -1 + \log(\rho) \nonumber \\
     & + (\rho \log(\rho+(e-1)))^{-1} \log(\log(\rho+(e-1))) 
\end{align}

Then, we shall prove the following lemma:
\begin{lemma}
$l'(\rho)$ has local minimum at $\rho = 1$ with $l(1) = 0$.  
\end{lemma}
(Proof)
Because $l''(\rho) > 0 $ for $\rho \ge 1$, $l'(\rho)$ is convex for $\rho \ge 1$. In addition, because $l'(1) = 1$, $l'(\rho)$ has a local minimum at $\rho = 1$ with $l(1) = 0$. Therefore, $l'(\rho) \ge 0 $ for $\rho \ge 1$. Since $l(0) = 0$, we can conclude that $l(\rho) \ge 0$ for $\rho \ge 1$. \qed

\end{document}